\pdfoutput=1
\documentclass[journal]{IEEEtran}

\usepackage{times}
\usepackage{epsfig}
\usepackage{graphicx}
\usepackage{amsmath}
\usepackage{amssymb}

\usepackage{cite}
\usepackage{float}
\usepackage{algorithm}
\usepackage{algorithmicx}
\usepackage{algpseudocode}
\usepackage{mathtools}
\usepackage{url}
\usepackage[table]{xcolor} 
\usepackage{threeparttable}
\usepackage{arydshln}
\usepackage[skip=0pt]{caption}
\usepackage{subcaption}
\usepackage{makecell}
\usepackage{pifont}
\usepackage{multirow}
\usepackage{booktabs}
\usepackage{chngcntr}

\usepackage{threeparttable}

\renewcommand{\vec}[1]{\mathbf{#1}}
\DeclareMathOperator*{\argmin}{arg\,min}

\usepackage[pagebackref=true,breaklinks=true,letterpaper=true,colorlinks,bookmarks=false]{hyperref}
\newcommand\blfootnote[1]{%
  \begingroup
  \renewcommand\thefootnote{}\footnote{#1}%
  \addtocounter{footnote}{-1}%
  \endgroup
}

\begin{document}

\title{Ground Plane Polling for 6DoF Pose Estimation of Objects on the Road}

\author{Akshay Rangesh and Mohan M. Trivedi\\
Laboratory for Intelligent \& Safe Automobiles, UC San Diego\\
{\tt\small \{arangesh, mtrivedi\}@ucsd.edu}
}

\maketitle
%
%
%
%

\blfootnote{Code: \href{https://github.com/arangesh/Ground-Plane-Polling}{https://github.com/arangesh/Ground-Plane-Polling}}

\begin{abstract}
This paper introduces an approach to produce accurate 3D detection boxes for objects on the ground using single monocular images. We do so by merging 2D visual cues, 3D object dimensions, and ground plane constraints to produce boxes that are robust against small errors and incorrect predictions. First, we train a single-shot convolutional neural network (CNN) that produces multiple visual and geometric cues of interest: 2D bounding boxes, 2D keypoints of interest, coarse object orientations and object dimensions. Subsets of these cues are then used to \textit{poll} probable ground planes from a pre-computed database of ground planes, to identify the ``best fit'' plane with highest consensus. Once identified, the ``best fit'' plane provides enough constraints to successfully construct the desired 3D detection box, without directly predicting the 6DoF pose of the object. 
The entire ground plane polling (GPP) procedure is constructed as a non-parametrized layer of the CNN that outputs the desired ``best fit'' plane and the corresponding 3D keypoints, which together define the final 3D bounding box. 
Doing so allows us to poll thousands of different ground plane configurations without adding considerable overhead, while also creating a single CNN that directly produces the desired output without the need for post processing. We evaluate our method on the 2D detection and orientation estimation benchmark from the challenging KITTI dataset, and provide additional comparisons for 3D metrics of importance. 
This single-stage, single-pass CNN results in superior localization and orientation estimation compared to more complex and computationally expensive monocular approaches.
\end{abstract}

\section{Introduction}\label{sec:introduction}

Localizing objects in 3D is of extreme importance in autonomous driving and driver safety applications. Traditional and contemporary approaches have mostly relied on range sensors like LiDARs and Radars, or stereo camera pairs to predict the desired 6DoF pose and dimensions of objects of interest. Some of these approaches are demonstrably robust under a variety of conditions, and produce high quality 3D detection boxes despite large occlusions, truncations etc. 
These approaches benefit from the use of 3D data, either as LiDAR point clouds, range measurements from Radars, depth maps obtained from stereo cameras, or a combination thereof. 
However, the benefits of 3D sensors are almost always accompanied by certain downsides. These sensors are typically orders of magnitudes more expensive compared to cheap cameras, and are also bulkier and power-hungry. It is therefore desirable to carry out 3D object detection with monocular cameras, if suitable robustness can be achieved. 
A robust 3D detector could also in turn improve the performance of purely camera based tracking~\cite{rangesh2018no}, prediction~\cite{deo2018would, dueholm2016trajectories, deo2018convolutional}, and other driver safety systems~\cite{deo2019control} and tools~\cite{zimmer20193d} in general. 
This however introduces many challenges, most of which stem from the fact that predicting 3D attributes from 2D measurements is an ill-posed problem. 


In this study, we overcome this challenge by only considering those cues (visual or otherwise), that can be reliably predicted using monocular camera images alone, while also being generalizable to data captured by cameras with different settings and parameters. With this in mind, we only predict attributes like 2D detection boxes, 2D keypoint locations, coarse local orientations, and dimensions of the object in 3D. State-of-the-art approaches for predicting 2D attributes like detection boxes and keypoints are known to generalize quite well to new datasets and scenarios. On the other hand, coarse local orientations are closely tied to the appearance of an object, and can be reliably predicted as we will show later. Finally, the dimensions of an object in 3D are somewhat tied to the appearance of an object (e.g. cars, vans, trucks etc. look different), but are less prone to large errors because of the low variance in dimensions of an object within a particular object class.

\begin{figure}[t]
\begin{center}
\includegraphics[width=0.9\linewidth]{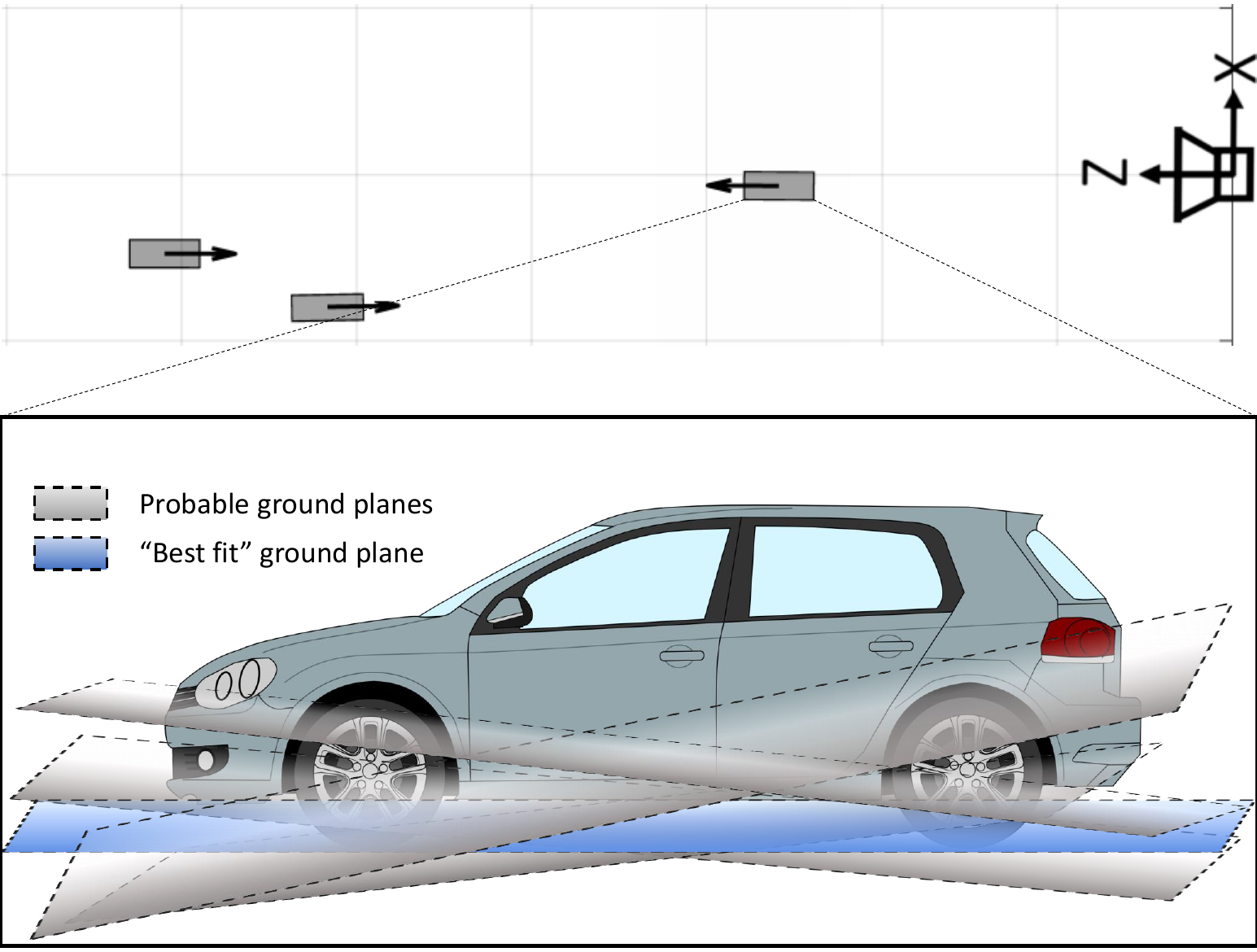}
\end{center}
\caption{Illustration of the ground plane constraint enforced in this study. Each object is assumed to lie on one of many probable ground plane configurations - termed to be the ``best fit'' plane associated with the object.}
\label{fig:motivation}
\end{figure}

\begin{figure*}[t]
\begin{center}
\includegraphics[width=\linewidth]{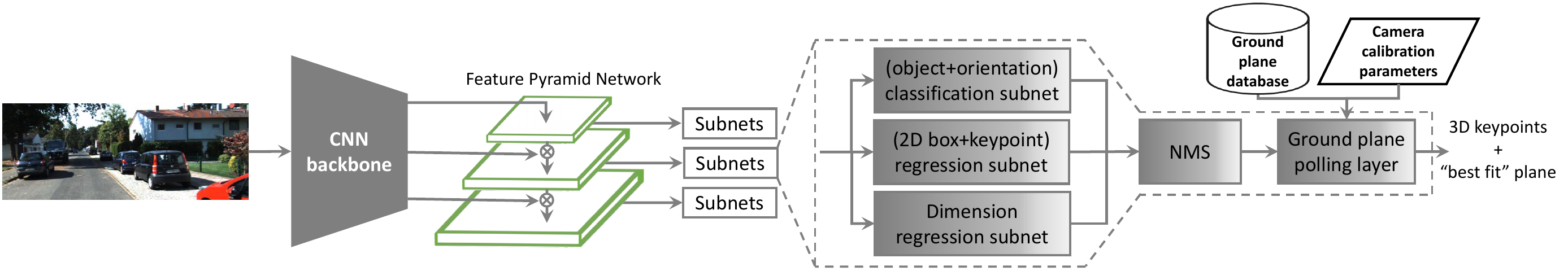}
\end{center}
\caption{Network architecture and our overall approach for predicting 3D bounding boxes from monocular images.}
\label{fig:network}
\vspace{-3mm}
\end{figure*}

On the other hand, we would like the constraints and assumptions we enforce to result in reasonable estimates, while not being too restrictive. Ground plane constraints are a common choice, where objects of interest are forced to lay on a common plane. This however, might be too restrictive of an assumption, resulting in large errors for objects farther away lying on irregular terrains. In this study, we instead create a database of probable ground planes, and choose the ``best fit'' plane for each object locally (see~Figure~\ref{fig:motivation}). This is similar to modeling the road terrain with a piecewise planar approximation, thereby loosening the single ground plane constraint. We also purposefully predict more attributes than needed to estimate a 3D detection box, and use these predictions to form maximal \textit{consensus set} of attributes, in a manner similar to traditional RANSAC approaches. This makes our approach robust to individual errors and outliers in predictions.

Our main contributions in this work can be summarized as follows - 1) We propose a single-shot approach to predict the 6DoF pose and dimensions of objects on the road by predicting 2D attributes of interest in single monocular images. 2)We then combine subsets of these attributes to robustly identify the ``best fit" ground plane for each object locally using a novel polling approach, thereby determining the 3D detection box corresponding to each object. 3) Finally, we carry out extensive comparisons and experiments with previous state-of-the-art techniques to illustrate the advantages and limitations of our approach.

\section{Related Research}

Since our approach is based on extending 2D object detectors to infer 6DoF pose, we first give a brief overview of recent 2D detection approaches. 2D detection of objects in single images has long been of interest to the computer vision community. Most recent works rely on convolutional neural networks (CNNs) to produce high quality 2D boxes for a variety of objects under challenging scenarios. 2D object detection is generally approached in two ways - using single shot networks, or by using multi-stage architectures. Single shot (or single stage) detectors like \cite{liu2016ssd, redmon2016you, lin2018focal} directly regress to the offset between predefined anchors in a grid, in a dense manner. Additionally, anchors are scored to determine which boxes to retain and the class of the corresponding object. Multi-stage detectors split these operations into two stages, namely the proposal generation stage, followed by the proposal processing stage\cite{ren2015faster, kong2016hypernet, yang2016exploit}. During inference, these detectors first propose candidate detection boxes, each of which is then individually processed and refined by a second network that outputs the desired 2D detection box and object class. Even though the object proposal stage reduces the search space significantly, having to process each proposal individually imposes a significant toll on the runtime. Consequently, two stage methods are generally slower than single shot methods, albeit with better performance. 

Classical methods for estimating 3D pose primarily involve identifying distinct keypoints and finding correspondences between different views by matching features. These feature descriptors were designed to be invariant to changes in scale, rotation, illumination and keypoints\cite{lowe1999object, rothganger20063d, rublee2011orb, bay2006surf}. Other related approaches involve 3D model-based registration\cite{li2011robustly, lowe1991fitting}, and Hausdorff and Chamfer matching for edges and curves\cite{huttenlocher1992comparing, liu2010fast, ramnath2014car}. Such methods are often fast and work reasonably well in cluttered scenes. However, they are too reliant on texture, high resolution images, and do not take into account high level information about the scene.

With the introduction of new large-scale datasets  like~\cite{Geiger2012CVPR, xiang2014beyond, matzen2013nyc3dcars} and overall success of CNNs, many approaches for 3D object detection and 6DoF pose estimation using single monocular images have recently emerged. Although these methods are mostly based on popular 2D detection architectures\cite{redmon2016you, lin2018focal, ren2015faster, yang2016exploit}, they differ in the ways they incorporate 3D information, the attributes they predict, and the constraints they enforce. 
In particular, we find that most monocular 3D detection methods can be categorized into the following three groups or a combination thereof - approaches that make use of 3D models, templates or exemplars, approaches that use 3D bounding box proposals, and approaches that enforce geometric constraints. We describe representative works from each category below.

Xiang et al.~\cite{xiang2015data} cluster the set of possible object poses into viewpoint-dependent subcategories using 3D voxel patterns. In their following work~\cite{xiang2017subcategory}, these subcategories were used as supervision to train a network to detect and classify the subcategory of each object. Subcategory information is then transferred to obtain the pose of each object. More recently, Chabot et al.~\cite{chabot2017deep} create a dataset of 3D shapes and templates, and identify the most similar template for each object that is detected. To do so, they manually annotate vehicle part coordinates, part visibility and the 3D template corresponding to each object in the training set. At test time, their network predicts all part coordinates, their visibility and the most similar template. A 2D/3D matching algorithm (Perspective-n-Point) then produces the desired object pose. These methods generally provide more information about each object, at the expense of having a more convoluted approach involving a database of shapes, templates, voxel patterns etc., and sometimes requiring more manual annotations.

Archetypal studies that make use of 3D object proposals for monocular detection are presented by Chen et al. in~\cite{chen20153d, chen2016monocular}. In these studies, the authors first sample candidate 3D boxes using a ground plane assumption and object size priors. These boxes are then scored by exploiting different cues like semantic and instance classes, context, shape, location etc. Although these methods work well on 2D detection tasks, they fail to localize objects accurately in 3D.

Unlike previous methods, some recent studies have chosen to enforce geometric constraints to obtain the 6DoF pose of objects. In~\cite{mousavian20173d}, Mousavian et al. only predict the orientation of objects, and use the  fact that the perspective projection of a 3D bounding box should fit tightly within its 2D detection window. 
This constraint, expressed as a linear system of equations, when solved, results in the desired 6DoF pose of the object. 
In a more straightforward approach, the authors in~\cite{tekin2017real} use a single shot network to predict 2D projections of all 8 corners and the centroid of the 3D bounding box, and use these 2D-3D correspondences to obtain the 6DoF pose of the object by solving the Perspective-n-Point (PnP) problem. These methods, although conceptually simple and straightforward to implement, rely on the accuracy of all predicted entities.

\section{Network Architecture}

\subsection{Overview}
Our primary focus in this study is to propose an approach to real time 3D object detection for the purpose of autonomous driving. With this in mind, our network design is based on the RetinaNet detector presented in~\cite{lin2018focal}. Although our approach can be adapted to work with any generic object detector based on anchor boxes, we decided to work with the RetinaNet architecture because it matches the speed of other one-stage detectors, while having comparable performance to state-of-the-art two stage detectors. As illustrated in Figure~\ref{fig:network}, we retain the backbone structure based on Feature Pyramid Networks~(FPN)~\cite{lin2017feature}, and modify and add to the subnetworks that follow. For most of our experiments, we use a ResNet50 backbone~\cite{he2016deep} with five pyramid levels ($P_3$ to $P_7$) computed using convolutional features $C_3$, $C_4$ and $C_5$ from the ResNet architecture as before. However, unlike~\cite{lin2018focal}, all pyramid levels have $C = 512$ channels instead of $256$ to account for the increase in the number of subnetwork outputs, and the number of channels per output. 
We describe the purpose and structure of each subnetwork in the subsections that follow. Details of the ground plane polling layer are presented in the following section.

Similar to previous works, we use translation-invariant anchor boxes with areas of $32^2$ to $512^2$ on pyramid levels $P_3$ to $P_7$ respectively. As is common practice, we use anchors at three aspect ratios (1:2, 1:1, 2:1). However, for denser scale coverage and to account for objects that are farther away, we consider 4 different relative scales ($2^{-1/3}, 2^0, 2^{1/3}, 2^{2/3}$) of each anchor aspect ratio, resulting in a total of $A = 12$ anchors per location, per level. These anchors together cover a scale range of $25-813$ pixels with respect to the network’s input image.

The multi-scale features obtained from each level of the FPN are then processed by three different subnetworks as shown in Figure~\ref{fig:network}. First, the (object + orientation) classification subnetwork outputs the object class and coarse 3D orientation range of each object in the scene. Next, the (2D box + keypoint) regression subnetwork regresses to the desired 2D box and keypoints of interest from designated anchor boxes with sufficient overlap. Finally, the dimension regression subnetwork outputs the dimensions (height, width, length) of the 3D box corresponding to objects at each anchor location. These subnetwork outputs are then passed on to a non-parametrized \textit{ground plane polling} layer. The ground plane polling layer takes in the outputs of each subnetwork, the camera calibration parameters, and a database of probable ground planes, and outputs the desired ``best fit" plane, and the corresponding 3D location of each keypoint. We describe the purpose and structure of each subnetwork in the subsections that follow. Details of the ground plane polling layer are presented in the following section.

\subsection{Subnetwork Architectures}

\begin{figure}[t]
\begin{center}
\includegraphics[width=0.95\linewidth]{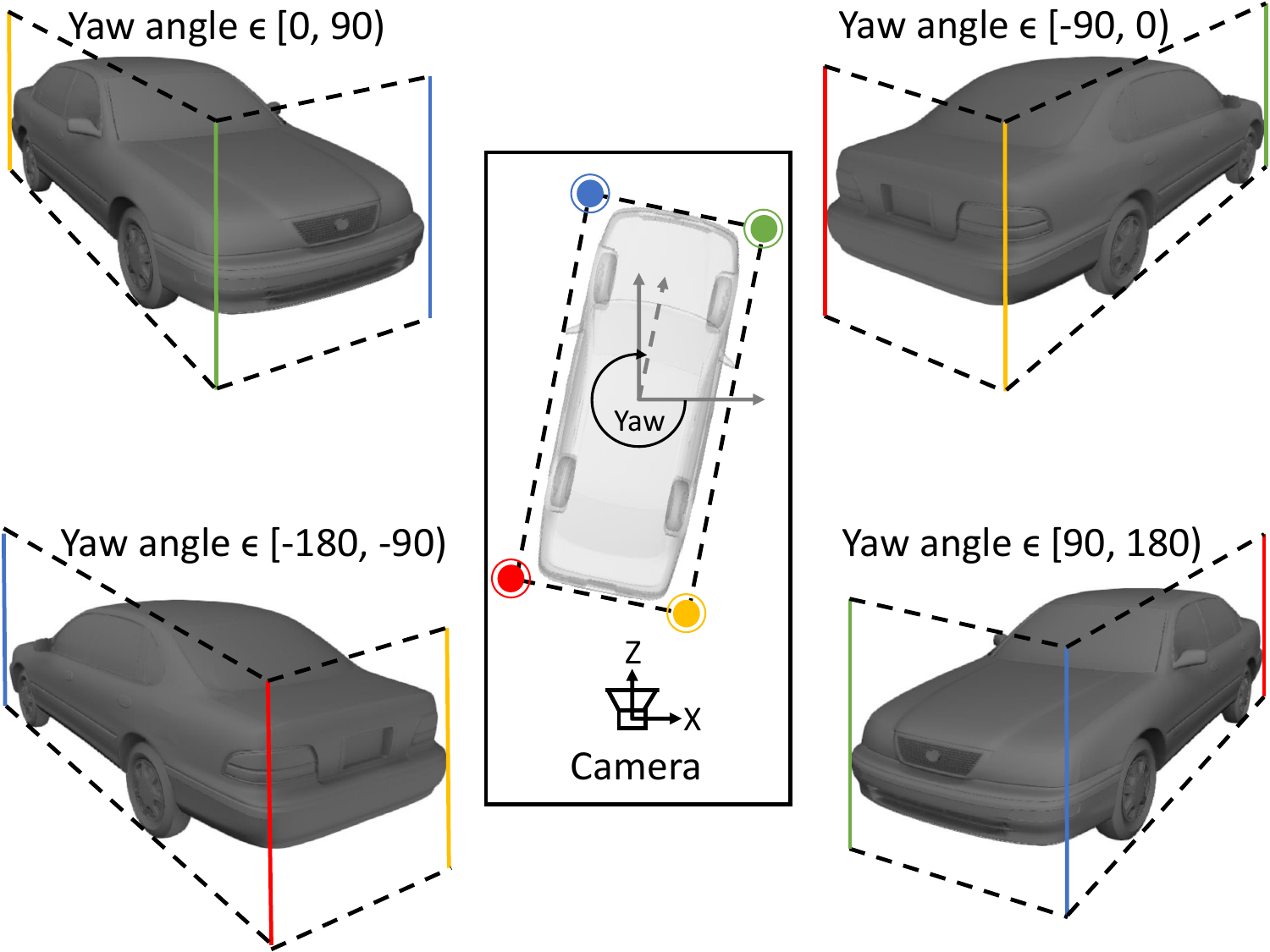}
\end{center}
\caption{Illustration of the 4 different orientation classes based on yaw angles (corners of the figure), and a bird's-eye view of how the yaw angle is calculated (center of the figure). Best viewed in color.}
\label{fig:orientations}
\end{figure}

\noindent\textbf{(object + orientation) classification subnet:} This subnetwork is similar to the classification head found in 2D detectors, with the notable addition of orientation classes. In particular, we define 4 coarse orientation classes tied to 4 different ranges into which an object's yaw angle in the camera coordinate frame may fall. These orientation classes and the corresponding change in object appearance is depicted in Figure~\ref{fig:orientations}. In this Figure, each vertical edge of the 3D bounding box is color coded to indicate its relationship to the orientation classes. 
Each orientation class is further split into two classes depending on the relative locations of certain keypoints of interest with respect to the center of each anchor. This yields a total of 8 orientation classes. We explain the nature of these split classes in our description of the regression subnetwork.

\begin{figure}[t]
\begin{center}
\includegraphics[width=0.7\linewidth]{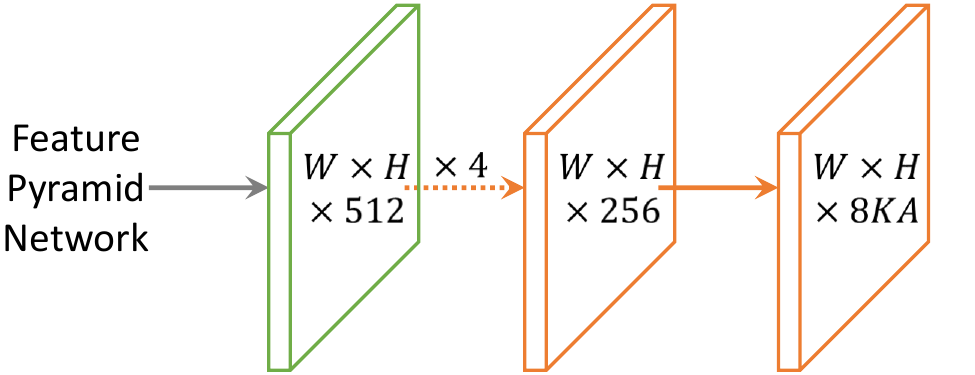}
\end{center}
\caption{\textbf{(object+orientation) classification subnet:} output channels accounts for 8 orientation classes, per anchor, per object class.}
\label{fig:classification_head}
\end{figure}

The classification subnetwork architecture is then modified to predict not just the object class, but also the desired orientation class. As can be seen in Figure~\ref{fig:classification_head}, this subnetwork takes in the output from each level of the FPN, applies a series of convolutional operations resulting in an output with $8KA$ channels to account for each of $8$ orientation classes, $K$ object classes, and $A$ anchors per location. We use a focal loss~\cite{lin2018focal} to train this subnetwork:
\begin{equation}\label{eq:focal}
L_{class}(\vec{p}, \vec{y}) = -\sum_{k=1}^{K} \sum_{o=1}^{8} \boldsymbol{\alpha}_{k, o} (1 - \vec{p}_{k, o})^\gamma \vec{y}_{k, o} \log(\vec{p}_{k, o}),
\end{equation}
where $\vec{p}$ and $\vec{y}$ are the estimated probabilities and one-hot ground truth vector corresponding to positive and negative anchors. 
Specifically, anchors are positive if they have an intersection-over-union (IoU) greater than $0.5$ with the ground truth and are considered negative if their IoU is in $[0, 0.4)$. Other anchors are ignored. $\gamma$ and $\boldsymbol{\alpha}$ are hyperparameters of the focal loss. While $\gamma$ is a simple scalar, $\boldsymbol{\alpha}$ represents the weights that account for foreground-background class imbalance and are defined as follows:
\begin{equation}
\boldsymbol{\alpha}_{k, o} = \begin{cases}
       \alpha, &\quad\text{if }\vec{y}_{k, o} = 1\\
       1 - \alpha, &\quad\text{otherwise}.\\
     \end{cases}
\end{equation}
The total classification loss is summed across all positive and negative anchors, and is normalized by the total number of positive anchors.

\noindent\textbf{(2D box + keypoint) regression subnet:} This subnetwork is used to regress to not only the desired 2D detection box, but also four additional keypoints of interest: $\vec{x}_l$, $\vec{x}_m$, $\vec{x}_r$, and $\vec{x}_t$ that denote the left, middle, right and top \textit{visible} corners of the desired 3D bounding box when projected into the image plane. We perform class and orientation agnostic regression for both the 2D boxes and the keypoints. This implies that a given keypoint may represent different corners of the 3D bounding box depending on the orientation class of the object in question. 

\begin{figure}[t]
\begin{center}
\includegraphics[width=0.95\linewidth]{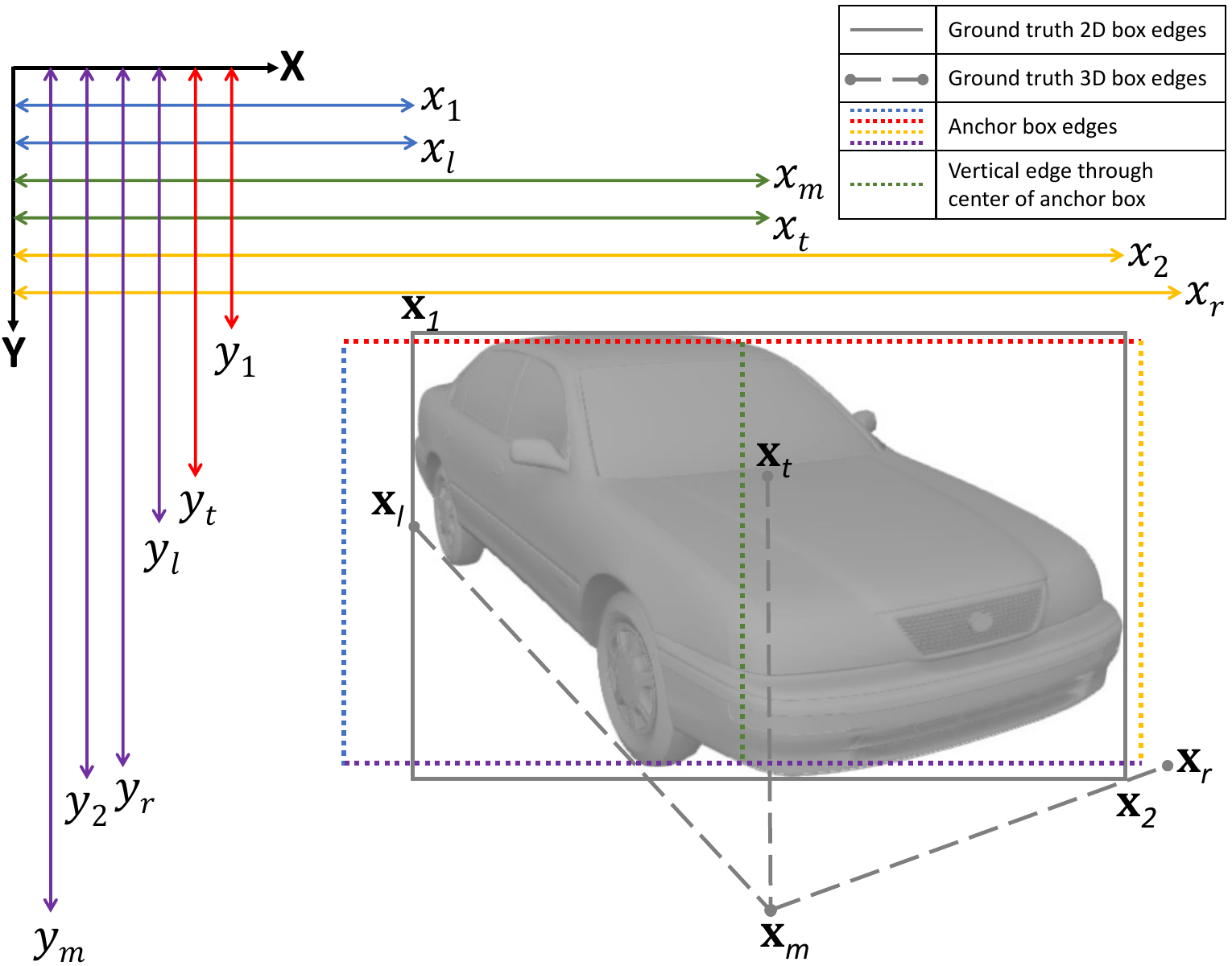}
\end{center}
\caption{Illustration of proposed regression targets for each anchor. Each target is colored based on the anchor edge from which it is regressed. Best viewed in color.}
\label{fig:anchors}
\end{figure}

\begin{figure}[t]
\begin{center}
\includegraphics[width=0.95\linewidth]{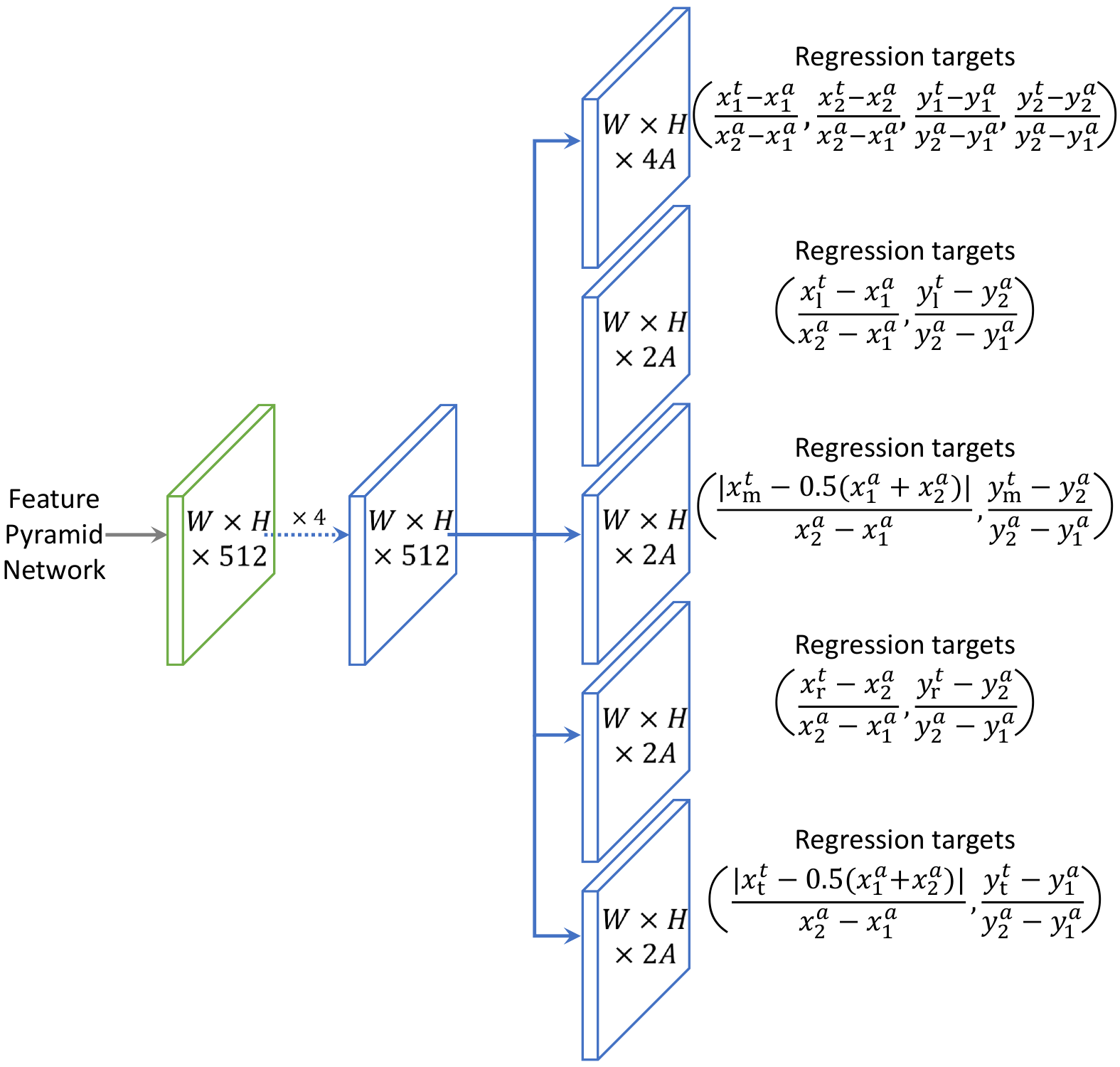}
\end{center}
\caption{\textbf{(2D box+keypoint) regression subnet:} we use separate heads for the 2D box and each keypoint. All regression targets associated with a target bounding box $(x_1^t, x_2^t, y_1^t, y_2^t)$, its keypoints $\{(x_l^t, y_l^t), (x_m^t, y_m^t), (x_r^t, y_r^t), (x_t^t, y_t^t)\}$, and a positive anchor $(x_1^a, x_2^a, y_1^a, y_2^a)$ are shown for reference.}
\label{fig:regression_head}
\end{figure}

Figure~\ref{fig:anchors} outlines the different regression targets for each positive anchor, and also the edge of the anchor from which each target is regressed. The subnetwork architecture and its outputs are depicted in Figure~\ref{fig:regression_head}.

For the $X$ coordinates $x_m$ and $x_t$ of keypoints $\vec{x}_m$ and $\vec{x}_t$, we only regress to the absolute values of the target (see Figure~\ref{fig:regression_head}). The corresponding signs of these regression targets for each anchor are accounted for by the split in orientation classes described in the classification subnetwork i.e. for each anchor, the classification network picks one of 8 orientation classes, which both defines the coarse orientation of the object, and the sign of the regression targets for $x_t$ and $x_m$. This is especially important for degenerate vehicle orientations close to the boundaries of different orientation classes. 
For example, an aligned vehicle directly in front of the camera could have keypoints $\vec{x}_m$ and $\vec{x}_t$ close to either the left or right edges of a positive anchor box depending on the orientation class that is chosen. 

We use a smooth $L1$ loss for all regression heads. The total loss for each positive anchor in this subnetwork is the sum of all individual losses:
\begin{equation}
\begin{split}
L_{reg}(\vec{\tilde{t}}, \vec{t}) = L_{smooth-L1}(\vec{\tilde{t}}_{2D}, \vec{t}_{2D})
+ L_{smooth-L1}(\vec{\tilde{t}}_{\vec{x}_l}, \vec{t}_{\vec{x}_l}) \\+ L_{smooth-L1}(\vec{\tilde{t}}_{\vec{x}_m}, \vec{t}_{\vec{x}_m})
+ L_{smooth-L1}(\vec{\tilde{t}}_{\vec{x}_r}, \vec{t}_{\vec{x}_r}) \\+ L_{smooth-L1}(\vec{\tilde{t}}_{\vec{x}_t}, \vec{t}_{\vec{x}_t}),\\
\end{split}
\end{equation}
where $\vec{\tilde{t}}$ and $\vec{t}$ are the regression outputs and the corresponding targets respectively. The total loss for this subnetwork is the average loss over all positive anchors.

\begin{figure}[t]
\begin{center}
\includegraphics[width=0.7\linewidth]{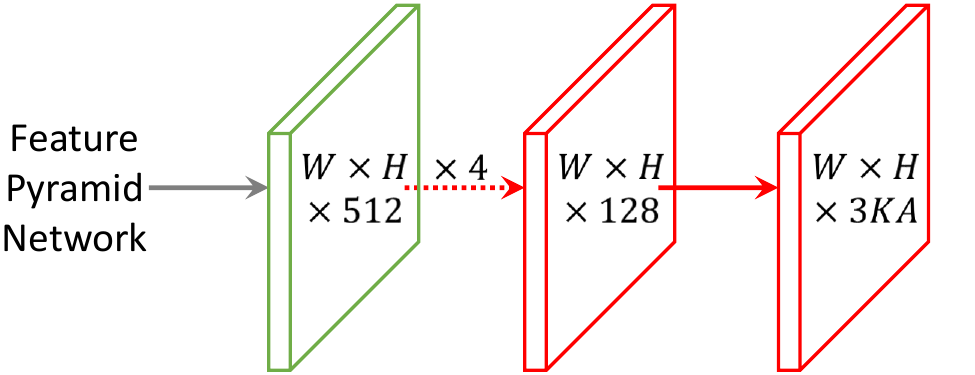}
\end{center}
\caption{\textbf{Dimension regression subnet:} output channels accounts for three dimensions (height, width, length), per anchor, per object class.}
\label{fig:dimension_head}
\end{figure}

\noindent\textbf{Dimension regression subnet:} This subnetwork has a similar architecture to the (object + orientation) classification subnetwork, but with fewer channels ($128$ instead of $256$) per convolution. The network directly outputs the three dimensions (height, width, length) of the desired 3D bounding box, but in a class specific manner. By making the predictions class specific, we improve the prediction accuracies just by reducing the variance within each class. The subnetwork architecture is shown in Figure~\ref{fig:dimension_head}, resulting in an output with $3KA$ channels, representing the 3 dimensions of the box, for each of $K$ object classes, and each of $A$ anchors per position.

The dimension regression subnetwork is trained using a smooth $L1$ loss: 
\begin{equation}
L_{dim}(\vec{\tilde{d}}, \vec{d}) = L_{smooth-L1}(\vec{\tilde{d}}, \vec{d}),
\end{equation}
where $\vec{\tilde{d}}$ and $\vec{d}$ are the regression outputs and the corresponding targets for all three dimensions respectively. The total loss for this subnetwork is average loss over all positive anchors.

The final loss for the entire network is a weighted sum of the three subnetwork losses:
\begin{equation}\label{eq:total}
L = L_{class} + \lambda_{reg} L_{reg} + \lambda_{dim} L_{dim}.
\end{equation}

\section{Ground Plane Identification by Polling}

\subsection{Database Creation}\label{sec:database}

As highlighted in Figure~\ref{fig:network}, the ground plane polling block takes in a database of probable ground planes and outputs the ``best fit" plane for each detection. This database of planes can be compiled either using heuristics based on the camera location, or by using 3D sensors like LiDARs to automatically fit planes to 3D data. In this study, we adopt the latter approach. In particular, we make use of the KITTI-15 dataset\cite{Alhaija2018IJCV} comprising of $200$ RGB images, LiDAR point clouds, and pixel-accurate semantic labels. 

Algorithm~\ref{alg:planes} describes the approach to creating a ground plane database using such data. The procedure involves identifying and retaining LiDAR points corresponding to semantic classes of interest, and then iteratively fitting planes to these points using RANSAC. Since we are interested in generating a large number of diverse ground planes, we use a very small inlier threshold ($t = 2cm$) and a very high probability of success ($p = 0.999$). This produces approximately $22$k ground plane candidates.

\begin{algorithm}[t]
\small
\caption{Pseudocode for creating a database of ground planes}\label{alg:planes}
\textbf{Inputs:} 
$\{im_i^{seg}$, \Comment{ground truth semantic segmentation}\\
\text{\hspace{1.2cm}} $\{(X_i^j, Y_i^j, Z_i^j)\}_{j=1}^M$, \Comment{LiDAR point cloud}\\
\text{\hspace{1.2cm}} $P_i\}_{i=1}^{N}$, \Comment{camera calibration matrix}\\
$semantic\_classes = \{ground, road, sidewalk, parking\}$\\
\text{\hfill} \Comment{semantic classes of interest}\\
\textbf{Outputs:} $\{\pi_k\}_{k=1}^K$ \Comment{database of ground planes}
\begin{algorithmic}[h]
\State $ground\_planes \gets \{\}$
\For{$i \gets 1$ \textbf{to} $N$}
\State $points \gets \{\}$
\For{$j \gets 1$ \textbf{to} $M$}
\State $(x_i^j, y_i^j) \gets project((X_i^j, Y_i^j, Z_i^j), P_i)$
\Statex \Comment{project LiDAR point to image plane}
\If{$im_i^{seg}(x_i^j, y_i^j) \in semantic\_classes$}
\State $points \gets points \bigcup \{(X_i^j, Y_i^j, Z_i^j)\}$
\EndIf
\EndFor
\While{$|points| \geq 3$}
\State $\pi, inliers \gets RANSAC(points)$
\State $ground\_planes \gets ground\_planes \bigcup \{\pi\}$
\State $points \gets points$\textbackslash $inliers$
\EndWhile
\EndFor
\State \textbf{return} $ground\_planes$
\end{algorithmic}
\end{algorithm}

\subsection{Ground Plane Polling}\label{sec:polling}

Given a ground plane and 2D keypoints $\vec{x}_l$, $\vec{x}_m$, and $\vec{x}_r$ of an object, it is straightforward to obtain the corresponding 3D keypoints lying on the plane. This is done by backprojecting a ray from the camera center through each keypoint, and finding its point of intersection with the plane. The backprojected ray $\vec{r} = (r_1, r_2, r_3, r_4)^T$ for a 2D keypoint $\vec{x} = (x, y)^T$ is created using the camera projection matrix,
\begin{equation}
\vec{r} = P^{+}[x, y, 1]^T,
\end{equation}
and the corresponding 3D keypoint on the plane $\pi = (a, b, c, d)^T$ is given by
\begin{equation}
\vec{X}^\pi = s[r_1, r_2, r_3]^T,
\end{equation}
where the scalar $s$ is defined as follows
\begin{equation}
s = \frac{-d r_4}{a r_1 + b r_2 + c r_3}.
\end{equation}
This results in the 3D keypoints $\vec{X}_l^\pi$, $\vec{X}_m^\pi$, $\vec{X}_r^\pi$ corresponding to 2D keypoints $\vec{x}_l$, $\vec{x}_m$, $\vec{x}_r$, lying on the plane $\pi$.

Unlike the other three keypoints, the desired point $\vec{X}_t^\pi$ corresponding to the 2D keypoint $\vec{x}_t$ does not lie on the plane $\pi$, but is rather a point on the line through $\vec{X}_m^\pi$ along the normal to the plane $\pi$. Since the desired 3D keypoint lies on both the backprojected ray $\vec{r}$ and the plane normal through $\vec{X}_m^\pi$, we could ideally calculate it from the intersection of these two lines. In practice, however, these are \textit{skew} lines that do not intersect. We therefore resort to an approximation, where $\vec{X}_t^\pi$ is designated as the point on the plane normal through $\vec{X}_m^\pi$ that is closest in distance to the ray $\vec{r}$. If $\vec{n}_\vec{r}$ and $\vec{n}_\pi$ are the unit vectors representing the directions of the ray $\vec{r}$ and the plane normal respectively, the desired point can be calculated as follows:
\begin{equation}
\vec{X}_t^\pi = \vec{X}_m^\pi - \frac{\vec{X}_m^\pi \cdot \big(\vec{n}_\vec{r} \times (\vec{n}_\pi \times \vec{n}_\vec{r})\big)}{\vec{n}_\pi \cdot \big(\vec{n}_\vec{r} \times (\vec{n}_\pi \times \vec{n}_\vec{r})\big)} \vec{n}_\pi.
\end{equation}

\begin{figure}[t]
\begin{center}
\includegraphics[width=0.7\linewidth]{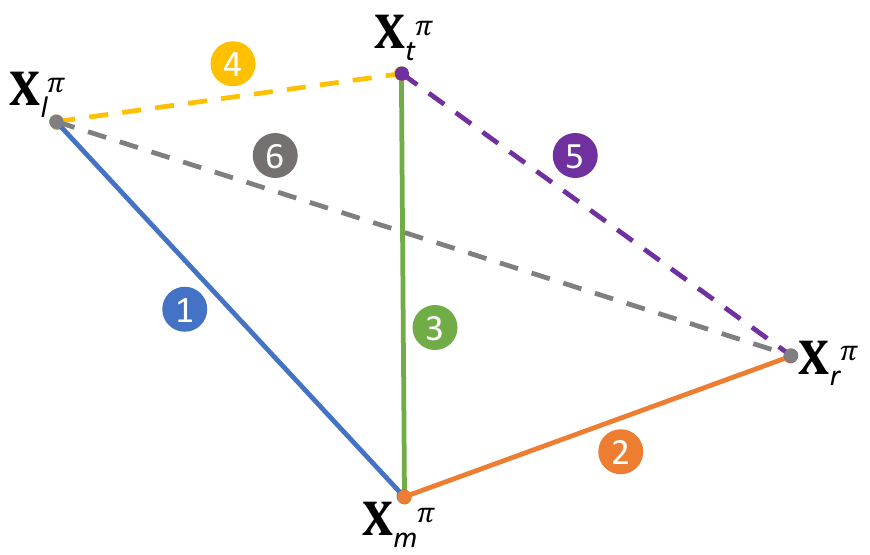}
\end{center}
\caption{Illustration of the $6$ unique combinations of 3D keypoints and the line segments joining them.}
\label{fig:triplets}
\end{figure}

To find out how well a ground plane $\pi$ fits the predicted 2D keypoints and 3D dimensions of an object, we propose an approach based on the construction of triplets comprised of two 3D keypoints, and the estimated length of the line segment joining them. Four 3D keypoints result in a total of $^4C_2 = 6$ unique pairs of keypoints, and therefore $6$ total triplets $\{(\vec{X}_i^\pi, \vec{X}_{i'}^\pi, l_i)\}_{i=1}^6$. 
Figure~\ref{fig:triplets} depicts these combinations. 
Note that the lengths of the line segments are determined using the 3D box dimensions predicted by the network.

\begin{figure*}[t]
\captionsetup[subfigure]{justification=centering}
   \centering
   \begin{subfigure}[t]{0.22\textwidth}
      \centering
      \includegraphics[width=\linewidth]{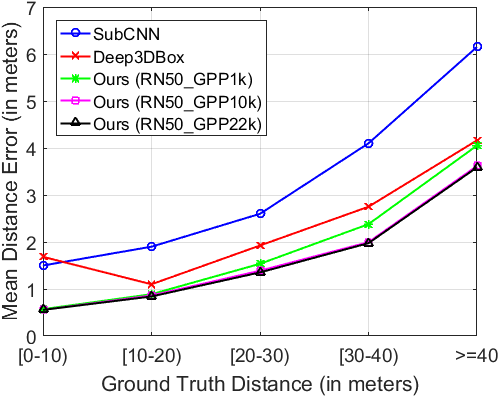}
      \caption{Mean error in estimating the 3D box center}
      \label{fig:us_vs_them_1}
   \end{subfigure}%
~     
   \begin{subfigure}[t]{0.22\textwidth}
      \centering
      \includegraphics[width=\linewidth]{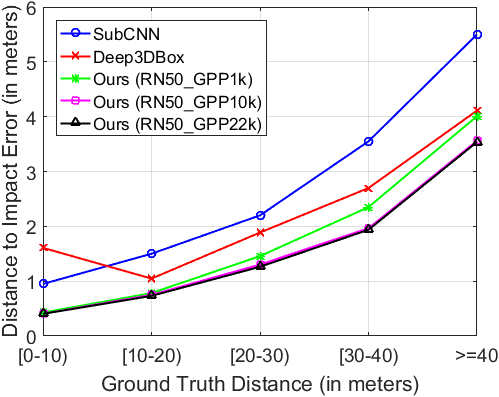}
      \caption{Mean error in estimating the closest point of the 3D box}
      \label{fig:us_vs_them_2}
   \end{subfigure}
~
   \begin{subfigure}[t]{0.22\textwidth}
      \centering
      \includegraphics[width=\linewidth]{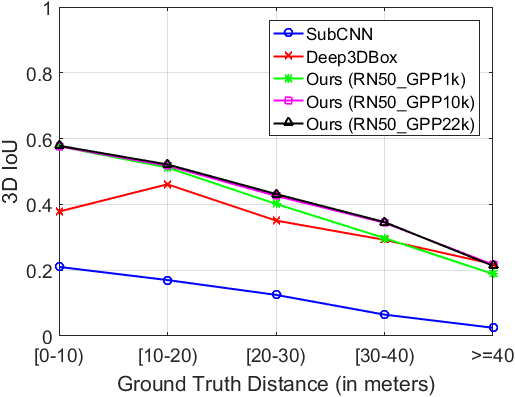}
      \caption{3D IoU between the predicted and ground truth 3D boxes}
      \label{fig:us_vs_them_3}
   \end{subfigure}%
~
   \begin{subfigure}[t]{0.25\textwidth}
      \centering
      \includegraphics[width=\linewidth]{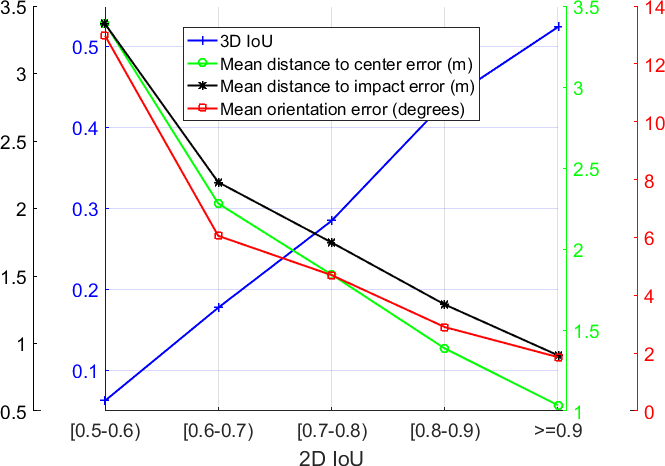}
      \caption{3D metrics as a function of 2D IoU for our approach.}
      \label{fig:2d_vs_3d}
   \end{subfigure}%
   \caption{Experiments related to 3D metrics of interest for different approaches on KITTI cars using the validation set provided in \cite{xiang2017subcategory, mousavian20173d}.}
  \vspace{-3mm}
\end{figure*}

Each triplet $(\vec{X}_i^\pi, \vec{X}_{i'}^\pi, l_i)$ is associated with a residual error defined by:
\begin{equation}
e_i^\pi = \big| ||\vec{X}_i^\pi - \vec{X}_{i'}^\pi|| - l_i \big|,
\end{equation}
and the total residual error $e^{\pi}$ for a plane is the sum of the six individual errors. Finally, the ``best fit" ground plane for a given object from a database of probable ground planes $\{\pi_k\}_{k=1}^K$ is determined to be
\vspace{-3mm}
\begin{equation}
\pi^* = \argmin_{\pi_k \in \{\pi_k\}_{k=1}^K} \sum_{i=1}^6 e_i^{\pi_k}.
\end{equation}
The residual errors tied to the three dimensions of the bounding box ensure that a plane that produces a box of reasonable dimensions is chosen. The other three errors corresponding to the face diagonals of the bounding box promote planes that result in boxes with nearly orthogonal edges. In our experiments, directly enforcing orthogonality or near-orthogonality led to most probable planes being discarded, thereby causing performance degradation. 

Knowing the ``best fit" plane $\pi^*$, we first discard one of the two keypoints $(\vec{X}_l^{\pi^*}$, $\vec{X}_r^{\pi^*})$ corresponding to the width of the bounding box. This is easily discerned from the predicted coarse orientation of the object. Doing so allows us to ensure orthogonality between adjacent sides, while retaining the predicted orientation (yaw). 
Next, we construct the desired 3D cuboid by utilizing $\vec{X}_l^{\pi^*}$ and the retained keypoint, the estimated 3D dimensions and the predicted orientation class. 
To account for a large database of ground planes, potentially large number of objects per image, and to leverage the GPU for matrix multiplications, we implement the entire polling procedure as a layer in our network. This is referred to as the ground plane polling (GPP) layer.

\section{Experimental Evaluation}

\subsection{Implementation Details}


\noindent\textbf{Training}: We use $\gamma = 2$ and $\alpha = 0.25$ for the focal loss presented in Equation~\ref{eq:focal}. The backbone of our network is pre-trained on ImageNet1k. The total loss depicted in Equation~\ref{eq:total} is fairly robust to the choice of hyperparamters $\lambda_{reg}$ and $\lambda_{dim}$ provided the network is trained for enough epochs. We set $\lambda_{reg} = \lambda_{dim} = 1$ for simplicity. The entire network is trained using mini-batch gradient descent using an Adam optimizer with learning rate $0.00001$, $\beta_1=0.9$ and $\beta_2=0.999$. We use a batch size of two images and train for a total of $70$ epochs on a single GPU. 
We apply random rotations in the image plane, translations, shears, scalings and horizontal flips to input images for robustness against small geometric transformations. 
Care was taken to apply the same transformations to all keypoints and also change the corresponding orientation class of objects when necessary. 
Additionally, the brightness, contrast, saturation and hue of input images were randomly perturbed. 
We noticed that data augmentation greatly helped with the robustness of our keypoint predictions.\\

\noindent\textbf{Inference}: During inference, we add the NMS (non-maximum suppression) and GPP layers to the end of the subnetwork outputs at each level of the FPN. To ensure fast operation, we only decode box predictions from at most $1$k top-scoring predictions per FPN level, after thresholding detector confidence at $0.05$. The top predictions from all levels are merged and non-maximum suppression with a threshold of $0.5$ is applied to yield the final 2D detections. These detections and their corresponding keypoints, dimensions and orientations are passed on to the GPP layer, which outputs the 3D keypoints and ``best fit" ground plane for each detection. The desired 3D bounding box is then constructed using these outputs.

\subsection{Results}

\begin{figure*}[t]
\captionsetup[subfigure]{justification=centering}
   \centering
   \begin{subfigure}[t]{0.23\textwidth}
      \centering
      \includegraphics[width=\linewidth]{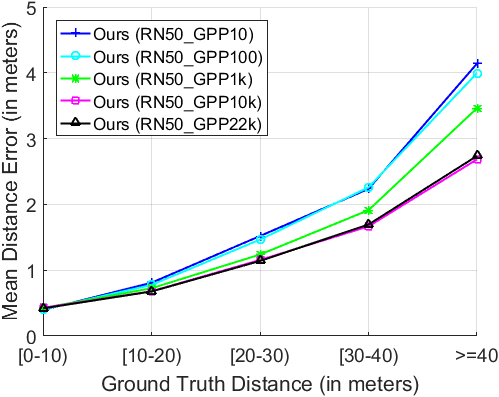}
      \caption{Mean error in estimating the 3D box center}
   \end{subfigure}%
~     
   \begin{subfigure}[t]{0.23\textwidth}
      \centering
      \includegraphics[width=\linewidth]{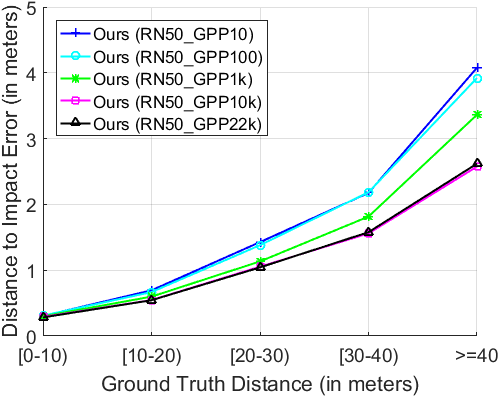}
      \caption{Mean error in estimating the closest point of the 3D box}
   \end{subfigure}
~     
   \begin{subfigure}[t]{0.23\textwidth}
      \centering
      \includegraphics[width=\linewidth]{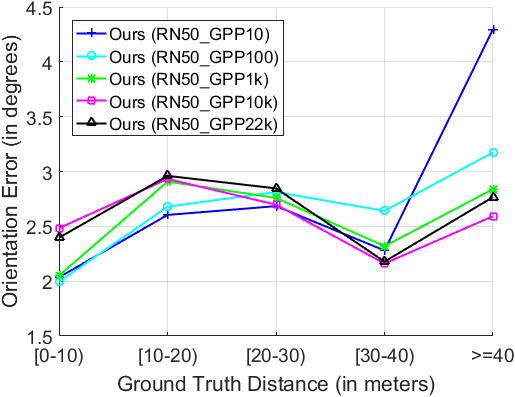}
      \caption{Mean orientation (yaw) error}
   \end{subfigure}
~
   \begin{subfigure}[t]{0.23\textwidth}
      \centering
      \includegraphics[width=\linewidth]{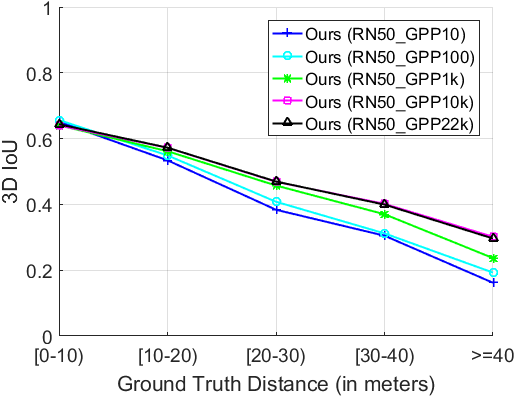}
      \caption{3D IoU between predicted and ground truth 3D boxes}
   \end{subfigure}%
   \caption{Effect of ground plane database size on 3D bounding box metrics for KITTI cars on a common validation set.}
   \label{fig:num_planes}
\end{figure*}

\begin{table*}[t]
\centering
\caption{\textbf{Results on the KITTI benchmark:}  Comparison of the Average Orientation Similarity (AOS), Average Precision (AP) and Orientation Score (OS) for cars. Our runtimes are reported using a RTX 2080 GPU.}
\resizebox{0.85\linewidth}{!}{%
\begin{threeparttable}
 \begin{tabular}{| c | c | c c c | c c c | c c c |}
 \hline
 \thead{Method} & \thead{Runtime} & \thead{} & \thead{Easy} & \thead{} & \thead{} & \thead{Moderate} &      \thead{} & \thead{} & \thead{Hard} & \thead{}\\
 \thead{} & \thead{(seconds)} & \thead{AOS} & \thead{AP} & \thead{OS} & \thead{AOS} & \thead{AP} & \thead{OS} & \thead{AOS} & \thead{AP} & \thead{OS}\\
 \hline \hline
  DeepMANTA\cite{chabot2017deep}\tnote{1} & 2.00 & 96.32\% & 96.40\% & 0.9991 & 89.91\% & 90.10\% & 0.9979 & 80.55\% & 80.79\% & 0.9970\\
  \hline
  Mono3D\cite{chen2016monocular} & 4.20 & 91.01\% & 92.33\% & 0.9984 & 86.62\% & 88.66\% & 0.9769 & 76.84\% & 78.96\% & 0.9731\\
  SubCNN\cite{xiang2017subcategory} & 2.00 & 90.67\% & 90.81\% & 0.9984 & 88.62\% & 89.04\% & 0.9952 & 78.68\% & 79.27\% & 0.9925\\
  Deep3DBox\cite{mousavian20173d} & 1.50 & 92.90\% & 92.98\% & 0.9991 & 88.75\% & 89.04\% & 0.9967 & 76.76\% & 77.17\% & \textbf{0.9947}\\
  Shift R-CNN\cite{naiden2019shift} & 0.25 & 90.27\% & 90.56\% & 0.9968 & 87.91\% & 88.90\% & 0.9889 & 78.72\% & 79.86\% & 0.9857\\
  MonoPSR\cite{ku2019monocular} & 0.20 & 89.88\% & 90.18\% & 0.9967 & 87.83\% & 88.84\% & 0.9886 & 70.48\% & 71.44\% & 0.9866\\
  \hdashline
    Ours (RN50\_GPP10k) & 0.21 & 89.42\% & 89.61\% & 0.9979 & 86.08\% & 87.02\% & 0.9892 & 76.47\% & 77.62\% & 0.9852\\
  Ours (RN101\_GPP10k) & 0.24 & 89.67\% & 89.84\% & 0.9981 & 86.63\% & 87.52\% & 0.9898 & 77.20\% & 78.36\% & 0.9852\\
  Ours (VGG16\_GPP10k) & 0.19 & 89.17\% & 89.26\% & 0.9990 & 87.16\% & 87.56\% & 0.9954 & 77.65\% & 78.29\% & 0.9918\\
  Ours (VGG19\_GPP10k) & 0.23 & 90.35\% & 90.42\% & \textbf{0.9992} & 87.96\% & 88.23\% & \textbf{0.9969} & 78.57\% & 79.00\% & 0.9946\\
  \hdashline
  Ours (VGG19\_GPP1k) & 0.11 & 90.12\% & 90.22\% & 0.9989 & 87.69\% & 88.06\% & 0.9958 & 78.41\% & 78.95\% & 0.9932\\
 \hline
 Ours (RN50\_FAST) & \textbf{0.05} & 88.86\% & 89.13\% & 0.9970 & 84.53\% & 85.67\% & 0.9867 & 75.18\% & 76.61\% & 0.9813\\
 Ours (VGG19\_FAST) & 0.07 & 89.87\% & 90.08\% & 0.9977 & 87.01\% & 87.59\% & 0.9934 & 77.59\% & 78.31\% & 0.9908\\
 \hline
 \end{tabular}
 \begin{tablenotes}
    \item[1] Uses additional keypoint labels and 3D CAD models not available to other methods.
  \end{tablenotes}
 \end{threeparttable}
 }
 \label{table:kitti}
 \vspace{-3mm}
\end{table*}

\begin{figure*}[h]
\begin{center}
\includegraphics[width=0.9\linewidth]{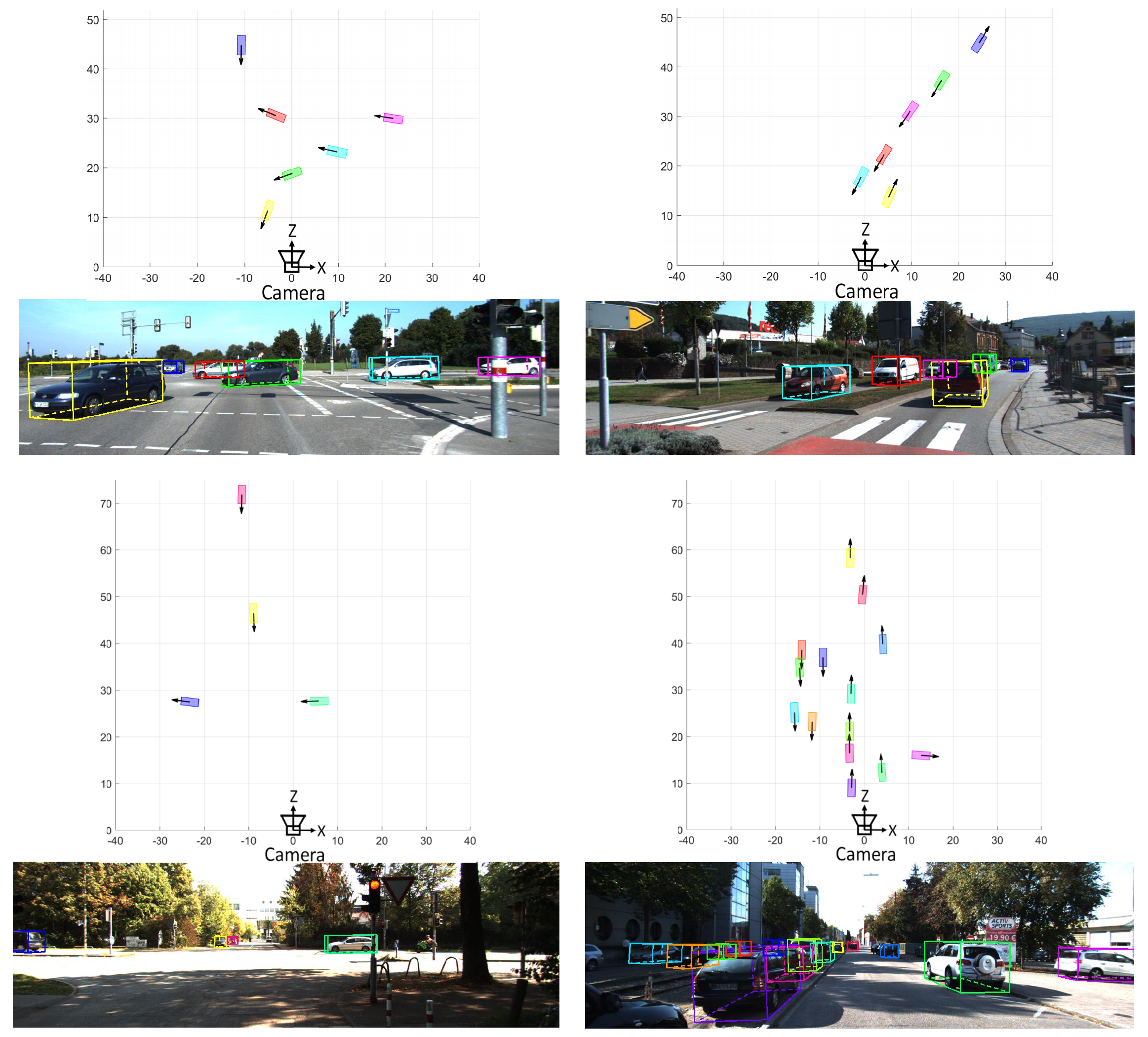}
\end{center}
\caption{\textbf{Qualitative results on the KITTI test set (VGG19\_GPP10k):} We show the bird's eye view of the predicted boxes (top), and their corresponding projections onto the input image (bottom) for four different scenes.}
\label{fig:kitti1}
\vspace{-3mm}
\end{figure*}

\noindent\textbf{Evaluation of 3D bounding box metrics}: The KITTI detection and orientation estimation benchmark~\cite{geiger2012we} for cars only evaluates the partial pose of vehicles. To provide a more complete evaluation and comparison with other monocular methods, we carry out analogous experiments to the ones proposed in \cite{mousavian20173d}. In particular, we plot three metrics of interest as a function of the distance of an object from the camera. The first metric is the average error in estimating the 3D coordinate of the center of objects. The second metric is the average error in estimating the closest point of the 3D bounding box from the camera. 
This metric is important for driving scenarios where the system needs to avoid hitting obstacles and is closely related to the \textit{time to collision} metric. 
The last metric is the standard 3D intersection over union (3D IoU) that depends on all factors of the 3D bounding box. 

In keeping with the original experiment, we factor away the 2D detection performance by considering only those detections that result in an IoU $\geq 0.7$ with the ground truth 2D box. This removes the effects of 2D detection performance and enables a side-by-side comparison of how well each method performs on the desired 3D metrics. To make the comparison fair, we use the same \textit{train-val} split provided in \cite{xiang2017subcategory}, where the KITTI training dataset comprising of $7481$ images is split into \textit{train} and \textit{val} sets consisting of $3619$ and $3799$ images respectively. 
Each method is trained on the \textit{train} set and is evaluated on the \textit{val} set. 
We compare our method to current (Deep3DBox~\cite{mousavian20173d}) and previous (SubCNN~\cite{xiang2017subcategory}) state-of-the-art monocular methods with publicly available results and/or code. We compare three variants of our method: RN50\_GPP1k, RN50\_GPP10k and RN50\_GPP22k, all three of which have a ResNet50 backbone, and differ only by the size of the ground plane database used in the GPP layer ($1$k, $10$k and $22$k planes respectively). As Figures~\ref{fig:us_vs_them_1}, \ref{fig:us_vs_them_2} and \ref{fig:us_vs_them_3} show, all three variants of our method outperform the other two methods on nearly all data points in the plots despite being an order of magnitude faster. In addition to these comparative results, we also plot the 3D metrics of interest as a function of 2D IoU for our model in Figure~\ref{fig:2d_vs_3d}. As 2D IoU with the ground truth increases, we observe monotonic improvements in all 3D metrics. This implies that a better 2D detector could result in better 3D pose.


\begin{figure*}[ht]
\begin{center}
\includegraphics[width=0.9\linewidth]{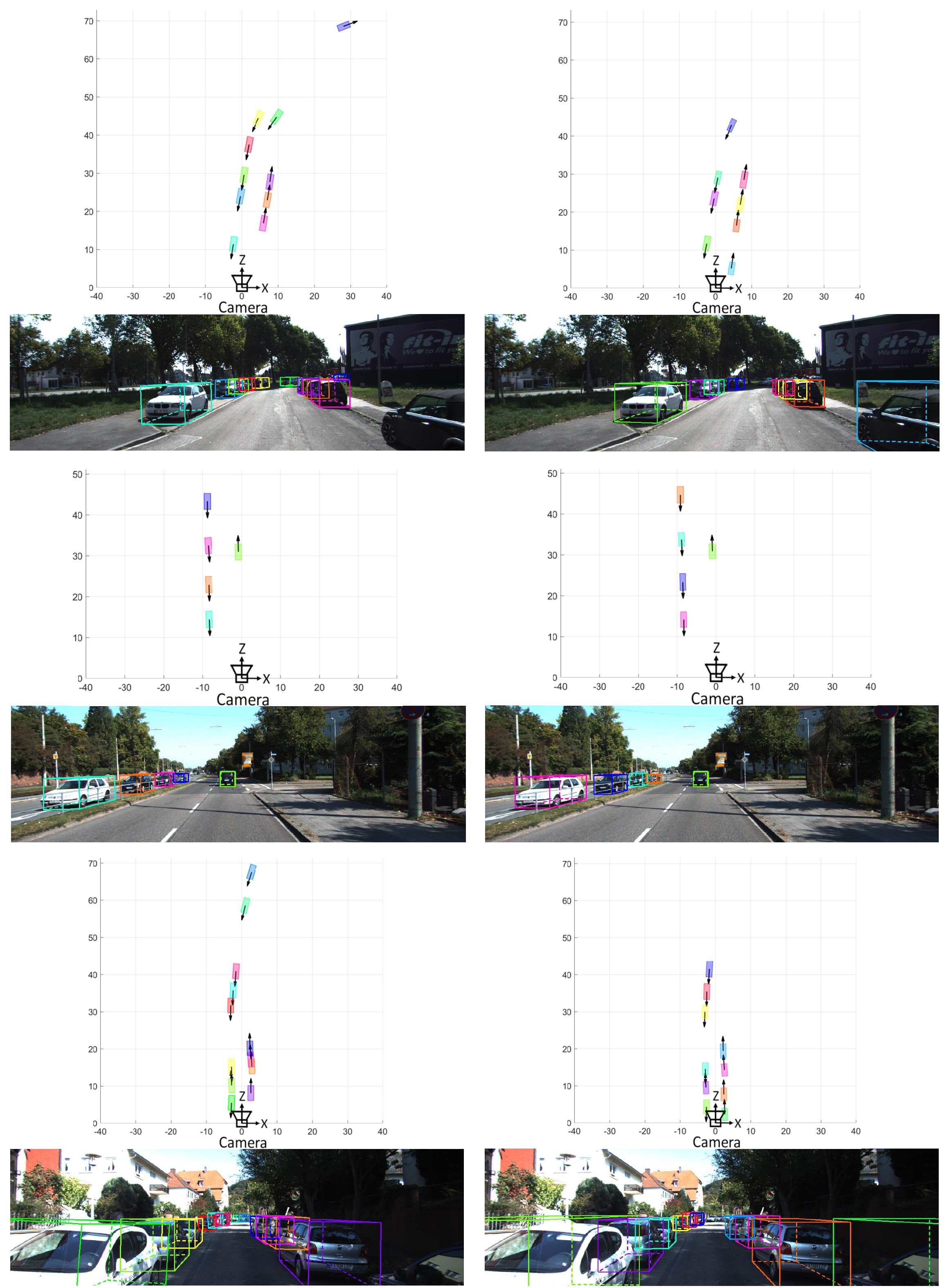}
\end{center}
\caption{\textbf{Qualitative results on the validation set (VGG19\_GPP10k):} We show the bird's eye view of the boxes (top), and their corresponding projections onto the input image (bottom) for both our model's predictions (left column) and the ground truth (right column).}
\label{fig:kitti2}
\vspace{-3mm}
\end{figure*}

\begin{figure*}[ht]
\begin{center}
\includegraphics[width=0.9\linewidth]{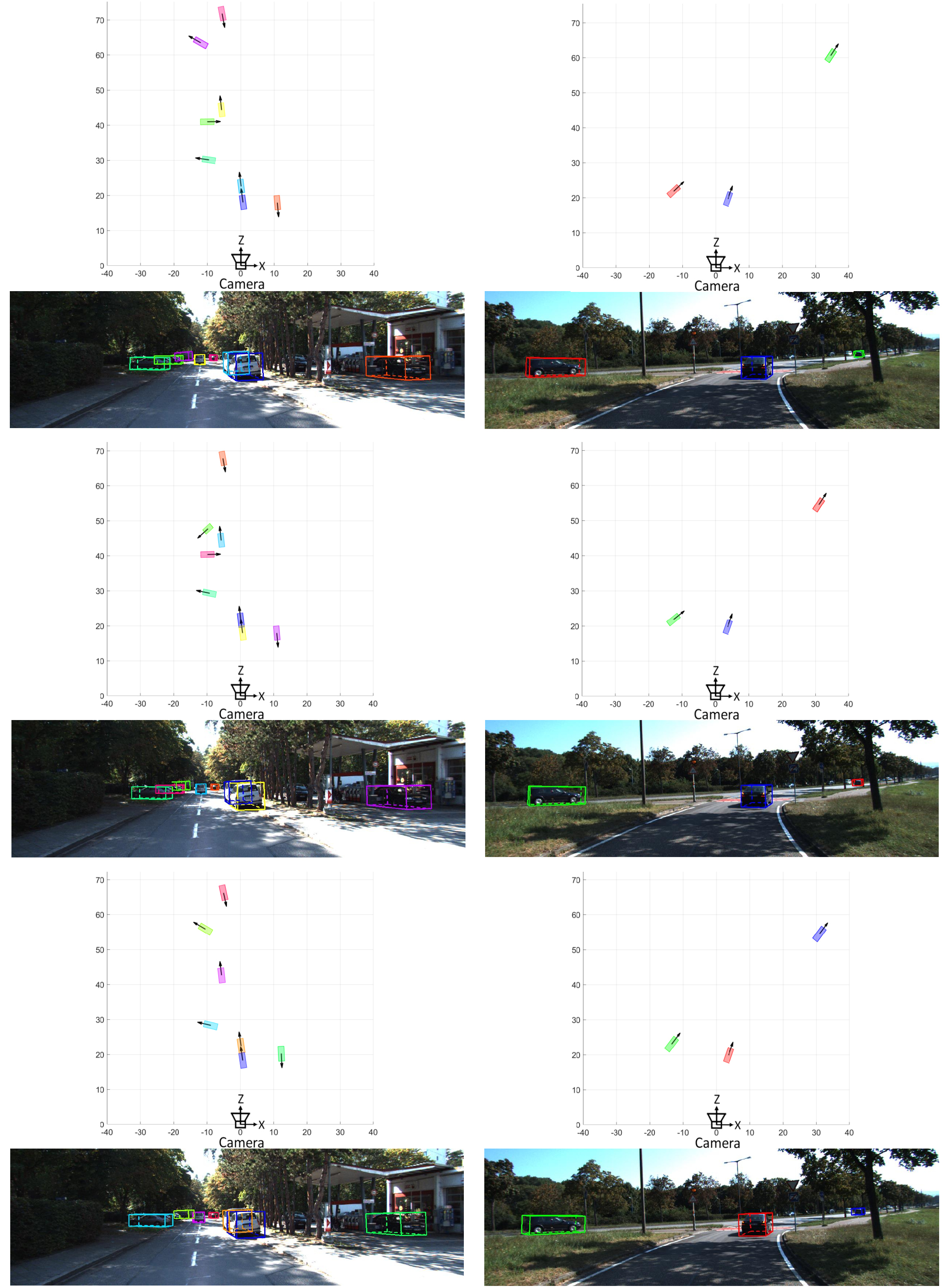}
\end{center}
\caption{\textbf{Qualitative results on the KITTI test set:} We show the bird's eye view of the predicted boxes (top), and their corresponding projections onto the input image (bottom) for three variants of our model: VGG19\_GPP10k (top row), VGG19\_GPP1k (middle row), and VGG19\_FAST (bottom row).}
\label{fig:kitti3}
\vspace{-3mm}
\end{figure*}

\noindent\textbf{Effect of ground plane database size}: To quantify the effect of the number of ground planes used in the database, we provide comparative analysis of different variants of our method using the same 3D metrics as before. Additionally, we also plot the orientation error (in degrees) averaged across all samples as a function of distance to the camera. Crucially, we create a different \textit{train-val} split to ensure that the network does not overfit, leading to a more accurate comparison. The resulting \textit{train} and \textit{val} sets consist of $6373$ and $1108$ images respectively. Each set is comprised of images from disparate drives with no overlap. 

As mentioned in Section~\ref{sec:database}, our complete database is made of $22$k candidate ground planes. To create databases of smaller sizes, we rank planes based on the number of inliers associated with them (see Algorithm~\ref{alg:planes}), and choose planes in a ``top-k'' fashion. While retaining the same ResNet50 backbone and subnetworks, we provide comparisons between databases of size $10$, $100$, $1$k, $10$k and $22$k. We refer to these 5 variants as  RN50\_GPP10, RN50\_GPP100, RN50\_GPP1k, RN50\_GPP10k and RN50\_GPP22k respectively.

Figure~\ref{fig:num_planes} depicts the comparison between these variants on all four metrics. 
As expected, we see an increase in performance over all metrics as the size of the ground plane database is increased. This effect is exacerbated as the distance of objects from the camera increases. The plots also seem to indicate that doubling the number of planes from $10$k to $22$k results in minor performance improvement, and even degradation in some cases. This hints at a case of diminishing returns beyond a database with $10$k ground planes. With this in mind, we use a database of $10$k ground planes as our default choice in other experiments.

\noindent\textbf{KITTI 2D detection and orientation benchmark}: The official 3D metric of
the KITTI dataset is Average Orientation Similarity (AOS), which is defined in~\cite{Geiger2012CVPR} and multiplies the average precision (AP) of the 2D detector with the average cosine distance similarity for the azimuth. Hence, AP is by definition the upper bound of AOS. Our results are summarized in Table~\ref{table:kitti}, along with other top entries on the KITTI leaderboard. Since the AP and AOS metrics are heavily influenced by the 2D detection performance, we additionally list the ratio of AOS over AP for each method as done in previous works. This ratio is representative of how each method performs only on orientation estimation, while factoring out the 2D detector performance. This score is referred to as the Orientation Score (OS), which represents the error $(1 + cos(\Delta\theta))/2$ averaged across all examples. 

A cursory glance at Table~\ref{table:kitti} indicates that our model is mostly within a few precision points of other methods in terms on AP and AOS. This is expected given that our method is the only single-stage approach that does not require object proposals of any kind, thereby resulting in a speedup not possible with other multi-stage methods. More importantly, our best performing model (VGG19\_GPP10k) only falls behind DeepMANTA\cite{chabot2017deep} by a small amount on the OS score, while beating out or remaining at par with other methods. We manage to do so at a fraction of the computational cost, and without requiring additional annotations associated with keypoints, part labels, and part visibility as needed in DeepMANTA. Additionally, our method is the only one that does not rely on computing additional features such as disparity, semantic segmentation, instance segmentation, point clouds etc., and does not need multi-stage processing as in \cite{chen2016monocular, chabot2017deep, mousavian20173d, naiden2019shift, ku2019monocular}.

\noindent\textbf{Effect of different backbones}: To observe the effect of different backbone sizes and architectures on the final result, we provide results for four popular backbone choices on the KITTI benchmark in Table~\ref{table:kitti}. We refer to these variants as RN50\_GPP10k, RN101\_GPP10k, VGG16\_GPP10k and VGG19\_GPP10k, each named after the corresponding backbone architecture used. 
The FPN for the RN101 (ResNet101) backbone~\cite{he2016deep} is constructed using the convolutional features $C_3$, $C_4$ and $C_5$, and the FPN for the two VGG architectures~\cite{simonyan2014very} are constructed using the features from the maxpooling layers $P_3$, $P_4$, and $P_5$ that follow the convolutional blocks $C_3$, $C_4$, and $C_5$ respectively.
Surprisingly, the smaller VGG backbones yield better results, especially for orientation-related metrics. The larger VGG19 backbone even results in superior AP across the board, and is therefore our best performing model. The RN101 backbone leads to a meager improvement in comparison to the smaller RN50 backbone, implying that adding more layers does not necessarily justify the return. We believe that the superior performance of the VGG variants is most likely explained by their use of smaller convolutional kernels, which preserves smaller details, resulting in better keypoint predictions. Consequently, better keypoint predictions result in better orientation estimates, and hence better orientation scores.\\ 

In addition to the backbones listed above, we also train a significantly smaller networks by halving the number of channels per layer in each subnet, and per pyramid level. Using a ResNet50 backbone and a database of $1$k ground planes, the RN50\_FAST network runs at 20 fps with only a small drop in overall performance (see Table~\ref{table:kitti}). 
VGG19\_FAST is constructed in the same manner as our RN50\_FAST variant, with the only difference being the backbone architecture. Similar to our full size models, the VGG19\_FAST variant outperforms the RN50\_FAST variant while incurring a small overhead. More importantly, this smaller model is on par or comparable in performance to full size variants RN50\_GPP10k and RN101\_GPP10k, while being considerably smaller and faster. These experiments further indicate the advantage of VGG backbones over ResNet backbones for the task at hand.

\noindent\textbf{Qualitative results on KITTI test set}: In addition to the quantitative results presented, we also show some qualitative results of our method on the KITTI test set in Figures~\ref{fig:kitti1}, \ref{fig:kitti2}, and \ref{fig:kitti3}.  All figures contain both the bird's eye view of all predicted boxes, and their corresponding projections into the image.

Figure~\ref{fig:kitti1} depicts exemplar results of our best performing model (VGG19\_GPP10k) on the KITTI test set. Figure~\ref{fig:kitti2} shows comparative results of the VGG19\_GPP10k variant with corresponding ground truth boxes on a validation set. Results indicate that our model produces boxes that largely agree with the ground truth in terms of location and orientation. Finally, Figure~\ref{fig:kitti3} compares the output of three different variants of our model: VGG19\_GPP10k, VGG19\_GPP1k, and VGG19\_FAST. The first two variants tend to produce similar results closer to the camera, while being inconsistent farther away from the camera. This observation concurs with our experiment on the effect of ground plane database size.

\section{Conclusion}
In this study, we introduce an approach to monocular 3D object detection by leveraging ground planes. This Ground Plane Polling (GPP) method works by merging 2D attributes like keypoints and coarse orientations with 3D information from probable ground plane configurations. By doing so, we also ensure that our network only predicts those entities that are known to generalize well across different conditions and datasets. Adapting to a new dataset would only involve reassessing the database of 3D ground planes. Additionally, our method produces a redundant set of cues and relies on identifying a suitable consensus set within, thereby resulting in robustness to individual errors and outliers. We have shown that our GPP approach outperforms other popular monocular approaches in terms of localization and orientation estimation, while remaining comparable to other methods in 2D detection performance, albeit with a significantly reduced inference time. Future work entails adopting our approach to recent two-stage detectors, and also conducting experiments to analyze its robustness to prediction errors and generalizability to more contemporary datasets like \cite{caesar2019nuscenes, rangesh2017multimodal}.


\bibliographystyle{IEEEtran}
\bibliography{ref}

\begin{thebibliography}{10}
\providecommand{\url}[1]{#1}
\csname url@samestyle\endcsname
\providecommand{\newblock}{\relax}
\providecommand{\bibinfo}[2]{#2}
\providecommand{\BIBentrySTDinterwordspacing}{\spaceskip=0pt\relax}
\providecommand{\BIBentryALTinterwordstretchfactor}{4}
\providecommand{\BIBentryALTinterwordspacing}{\spaceskip=\fontdimen2\font plus
\BIBentryALTinterwordstretchfactor\fontdimen3\font minus
  \fontdimen4\font\relax}
\providecommand{\BIBforeignlanguage}[2]{{%
\expandafter\ifx\csname l@#1\endcsname\relax
\typeout{** WARNING: IEEEtran.bst: No hyphenation pattern has been}%
\typeout{** loaded for the language `#1'. Using the pattern for}%
\typeout{** the default language instead.}%
\else
\language=\csname l@#1\endcsname
\fi
#2}}
\providecommand{\BIBdecl}{\relax}
\BIBdecl

\bibitem{rangesh2018no}
A.~Rangesh and M.~M. Trivedi, ``No blind spots: Full-surround multi-object
  tracking for autonomous vehicles using cameras \& lidars,'' \emph{arXiv
  preprint arXiv:1802.08755}, 2018.

\bibitem{deo2018would}
N.~Deo, A.~Rangesh, and M.~M. Trivedi, ``How would surround vehicles move? a
  unified framework for maneuver classification and motion prediction,''
  \emph{IEEE Transactions on Intelligent Vehicles}, vol.~3, no.~2, pp.
  129--140, 2018.

\bibitem{dueholm2016trajectories}
J.~V. Dueholm, M.~S. Kristoffersen, R.~K. Satzoda, T.~B. Moeslund, and M.~M.
  Trivedi, ``Trajectories and maneuvers of surrounding vehicles with panoramic
  camera arrays,'' \emph{IEEE Transactions on Intelligent Vehicles}, vol.~1,
  no.~2, pp. 203--214, 2016.

\bibitem{deo2018convolutional}
N.~Deo and M.~M. Trivedi, ``Convolutional social pooling for vehicle trajectory
  prediction,'' in \emph{Proceedings of the IEEE Conference on Computer Vision
  and Pattern Recognition Workshops}, 2018, pp. 1468--1476.

\bibitem{deo2019control}
N.~Deo, N.~Meoli, A.~Rangesh, and M.~Trivedi, ``On control transitions in
  autonomous driving: A framework and analysis for characterizing scene
  complexity,'' in \emph{Proceedings of the IEEE International Conference on
  Computer Vision Workshops}, 2019, pp. 0--0.

\bibitem{zimmer20193d}
W.~Zimmer, A.~Rangesh, and M.~Trivedi, ``3d bat: A semi-automatic, web-based 3d
  annotation toolbox for full-surround, multi-modal data streams,'' \emph{arXiv
  preprint arXiv:1905.00525}, 2019.

\bibitem{liu2016ssd}
W.~Liu, D.~Anguelov, D.~Erhan, C.~Szegedy, S.~Reed, C.-Y. Fu, and A.~C. Berg,
  ``Ssd: Single shot multibox detector,'' in \emph{European conference on
  computer vision}.\hskip 1em plus 0.5em minus 0.4em\relax Springer, 2016, pp.
  21--37.

\bibitem{redmon2016you}
J.~Redmon, S.~Divvala, R.~Girshick, and A.~Farhadi, ``You only look once:
  Unified, real-time object detection,'' in \emph{Proceedings of the IEEE
  conference on computer vision and pattern recognition}, 2016, pp. 779--788.

\bibitem{lin2018focal}
T.-Y. Lin, P.~Goyal, R.~Girshick, K.~He, and P.~Doll{\'a}r, ``Focal loss for
  dense object detection,'' \emph{IEEE transactions on pattern analysis and
  machine intelligence}, 2018.

\bibitem{ren2015faster}
S.~Ren, K.~He, R.~Girshick, and J.~Sun, ``Faster r-cnn: Towards real-time
  object detection with region proposal networks,'' in \emph{Advances in neural
  information processing systems}, 2015, pp. 91--99.

\bibitem{kong2016hypernet}
T.~Kong, A.~Yao, Y.~Chen, and F.~Sun, ``Hypernet: Towards accurate region
  proposal generation and joint object detection,'' in \emph{Proceedings of the
  IEEE conference on computer vision and pattern recognition}, 2016, pp.
  845--853.

\bibitem{yang2016exploit}
F.~Yang, W.~Choi, and Y.~Lin, ``Exploit all the layers: Fast and accurate cnn
  object detector with scale dependent pooling and cascaded rejection
  classifiers,'' in \emph{Proceedings of the IEEE conference on computer vision
  and pattern recognition}, 2016, pp. 2129--2137.

\bibitem{lowe1999object}
D.~G. Lowe \emph{et~al.}, ``Object recognition from local scale-invariant
  features.'' in \emph{iccv}, vol.~99, no.~2, 1999, pp. 1150--1157.

\bibitem{rothganger20063d}
F.~Rothganger, S.~Lazebnik, C.~Schmid, and J.~Ponce, ``3d object modeling and
  recognition using local affine-invariant image descriptors and multi-view
  spatial constraints,'' \emph{International Journal of Computer Vision},
  vol.~66, no.~3, pp. 231--259, 2006.

\bibitem{rublee2011orb}
E.~Rublee, V.~Rabaud, K.~Konolige, and G.~Bradski, ``Orb: An efficient
  alternative to sift or surf,'' 2011.

\bibitem{bay2006surf}
H.~Bay, T.~Tuytelaars, and L.~Van~Gool, ``Surf: Speeded up robust features,''
  in \emph{European conference on computer vision}.\hskip 1em plus 0.5em minus
  0.4em\relax Springer, 2006, pp. 404--417.

\bibitem{li2011robustly}
Y.~Li, L.~Gu, and T.~Kanade, ``Robustly aligning a shape model and its
  application to car alignment of unknown pose,'' \emph{IEEE transactions on
  pattern analysis and machine intelligence}, vol.~33, no.~9, pp. 1860--1876,
  2011.

\bibitem{lowe1991fitting}
D.~G. Lowe, ``Fitting parameterized three-dimensional models to images,''
  \emph{IEEE Transactions on Pattern Analysis \& Machine Intelligence}, no.~5,
  pp. 441--450, 1991.

\bibitem{huttenlocher1992comparing}
D.~P. Huttenlocher, W.~J. Rucklidge, and G.~A. Klanderman, ``Comparing images
  using the hausdorff distance under translation,'' in \emph{Proceedings 1992
  IEEE Computer Society Conference on Computer Vision and Pattern
  Recognition}.\hskip 1em plus 0.5em minus 0.4em\relax IEEE, 1992, pp.
  654--656.

\bibitem{liu2010fast}
M.-Y. Liu, O.~Tuzel, A.~Veeraraghavan, and R.~Chellappa, ``Fast directional
  chamfer matching,'' in \emph{2010 IEEE Computer Society Conference on
  Computer Vision and Pattern Recognition}.\hskip 1em plus 0.5em minus
  0.4em\relax IEEE, 2010, pp. 1696--1703.

\bibitem{ramnath2014car}
K.~Ramnath, S.~N. Sinha, R.~Szeliski, and E.~Hsiao, ``Car make and model
  recognition using 3d curve alignment,'' in \emph{IEEE Winter Conference on
  Applications of Computer Vision}.\hskip 1em plus 0.5em minus 0.4em\relax
  IEEE, 2014, pp. 285--292.

\bibitem{Geiger2012CVPR}
A.~Geiger, P.~Lenz, and R.~Urtasun, ``Are we ready for autonomous driving? the
  kitti vision benchmark suite,'' in \emph{Conference on Computer Vision and
  Pattern Recognition (CVPR)}, 2012.

\bibitem{xiang2014beyond}
Y.~Xiang, R.~Mottaghi, and S.~Savarese, ``Beyond pascal: A benchmark for 3d
  object detection in the wild,'' in \emph{Applications of Computer Vision
  (WACV), 2014 IEEE Winter Conference on}.\hskip 1em plus 0.5em minus
  0.4em\relax IEEE, 2014, pp. 75--82.

\bibitem{matzen2013nyc3dcars}
K.~Matzen and N.~Snavely, ``Nyc3dcars: A dataset of 3d vehicles in geographic
  context,'' in \emph{Proceedings of the IEEE International Conference on
  Computer Vision}, 2013, pp. 761--768.

\bibitem{xiang2015data}
Y.~Xiang, W.~Choi, Y.~Lin, and S.~Savarese, ``Data-driven 3d voxel patterns for
  object category recognition,'' in \emph{Proceedings of the IEEE Conference on
  Computer Vision and Pattern Recognition}, 2015, pp. 1903--1911.

\bibitem{xiang2017subcategory}
------, ``Subcategory-aware convolutional neural networks for object proposals
  and detection,'' in \emph{Applications of Computer Vision (WACV), 2017 IEEE
  Winter Conference on}.\hskip 1em plus 0.5em minus 0.4em\relax IEEE, 2017, pp.
  924--933.

\bibitem{chabot2017deep}
F.~Chabot, M.~Chaouch, J.~Rabarisoa, C.~Teuli{\`e}re, and T.~Chateau, ``Deep
  manta: A coarse-to-fine many-task network for joint 2d and 3d vehicle
  analysis from monocular image,'' in \emph{Proc. IEEE Conf. Comput. Vis.
  Pattern Recognit.(CVPR)}, 2017, pp. 2040--2049.

\bibitem{chen20153d}
X.~Chen, K.~Kundu, Y.~Zhu, A.~G. Berneshawi, H.~Ma, S.~Fidler, and R.~Urtasun,
  ``3d object proposals for accurate object class detection,'' in
  \emph{Advances in Neural Information Processing Systems}, 2015, pp. 424--432.

\bibitem{chen2016monocular}
X.~Chen, K.~Kundu, Z.~Zhang, H.~Ma, S.~Fidler, and R.~Urtasun, ``Monocular 3d
  object detection for autonomous driving,'' in \emph{Proceedings of the IEEE
  Conference on Computer Vision and Pattern Recognition}, 2016, pp. 2147--2156.

\bibitem{mousavian20173d}
A.~Mousavian, D.~Anguelov, J.~Flynn, and J.~Ko{\v{s}}eck{\'a}, ``3d bounding
  box estimation using deep learning and geometry,'' in \emph{Computer Vision
  and Pattern Recognition (CVPR), 2017 IEEE Conference on}.\hskip 1em plus
  0.5em minus 0.4em\relax IEEE, 2017, pp. 5632--5640.

\bibitem{tekin2017real}
B.~Tekin, S.~N. Sinha, and P.~Fua, ``Real-time seamless single shot 6d object
  pose prediction,'' \emph{arXiv preprint arXiv:1711.08848}, 2017.

\bibitem{lin2017feature}
T.-Y. Lin, P.~Doll{\'a}r, R.~B. Girshick, K.~He, B.~Hariharan, and S.~J.
  Belongie, ``Feature pyramid networks for object detection.'' in \emph{CVPR},
  vol.~1, no.~2, 2017, p.~4.

\bibitem{he2016deep}
K.~He, X.~Zhang, S.~Ren, and J.~Sun, ``Deep residual learning for image
  recognition,'' in \emph{Proceedings of the IEEE conference on computer vision
  and pattern recognition}, 2016, pp. 770--778.

\bibitem{Alhaija2018IJCV}
H.~Alhaija, S.~Mustikovela, L.~Mescheder, A.~Geiger, and C.~Rother, ``Augmented
  reality meets computer vision: Efficient data generation for urban driving
  scenes,'' \emph{International Journal of Computer Vision (IJCV)}, 2018.

\bibitem{naiden2019shift}
A.~Naiden, V.~Paunescu, G.~Kim, B.~Jeon, and M.~Leordeanu, ``Shift r-cnn: Deep
  monocular 3d object detection with closed-form geometric constraints,''
  \emph{arXiv preprint arXiv:1905.09970}, 2019.

\bibitem{ku2019monocular}
J.~Ku, A.~D. Pon, and S.~L. Waslander, ``Monocular 3d object detection
  leveraging accurate proposals and shape reconstruction,'' \emph{arXiv
  preprint arXiv:1904.01690}, 2019.

\bibitem{geiger2012we}
A.~Geiger, P.~Lenz, and R.~Urtasun, ``Are we ready for autonomous driving? the
  kitti vision benchmark suite,'' in \emph{2012 IEEE Conference on Computer
  Vision and Pattern Recognition}.\hskip 1em plus 0.5em minus 0.4em\relax IEEE,
  2012, pp. 3354--3361.

\bibitem{simonyan2014very}
K.~Simonyan and A.~Zisserman, ``Very deep convolutional networks for
  large-scale image recognition,'' \emph{arXiv preprint arXiv:1409.1556}, 2014.

\bibitem{caesar2019nuscenes}
H.~Caesar, V.~Bankiti, A.~H. Lang, S.~Vora, V.~E. Liong, Q.~Xu, A.~Krishnan,
  Y.~Pan, G.~Baldan, and O.~Beijbom, ``nuscenes: A multimodal dataset for
  autonomous driving,'' \emph{arXiv preprint arXiv:1903.11027}, 2019.

\bibitem{rangesh2017multimodal}
A.~Rangesh, K.~Yuen, R.~K. Satzoda, R.~N. Rajaram, P.~Gunaratne, and M.~M.
  Trivedi, ``A multimodal, full-surround vehicular testbed for naturalistic
  studies and benchmarking: Design, calibration and deployment,'' \emph{arXiv
  preprint arXiv:1709.07502}, 2017.

\end{thebibliography}

\begin{IEEEbiography}[{\includegraphics[width=0.96in,height=1.25in,clip,keepaspectratio]{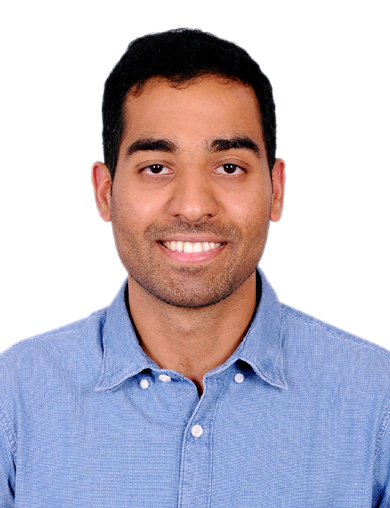}}]{Akshay Rangesh}
is currently working towards his PhD in electrical engineering from the University of California at San Diego (UCSD), with a focus on intelligent systems, robotics, and control. His research interests span computer vision and machine learning, with a focus on object detection and tracking, human activity recognition, and driver safety systems in general. He is also particularly interested in sensor fusion and multi-modal approaches for real time algorithms.
\end{IEEEbiography}

\begin{IEEEbiography}[{\includegraphics[width=1in,height=1.15in,clip,keepaspectratio]{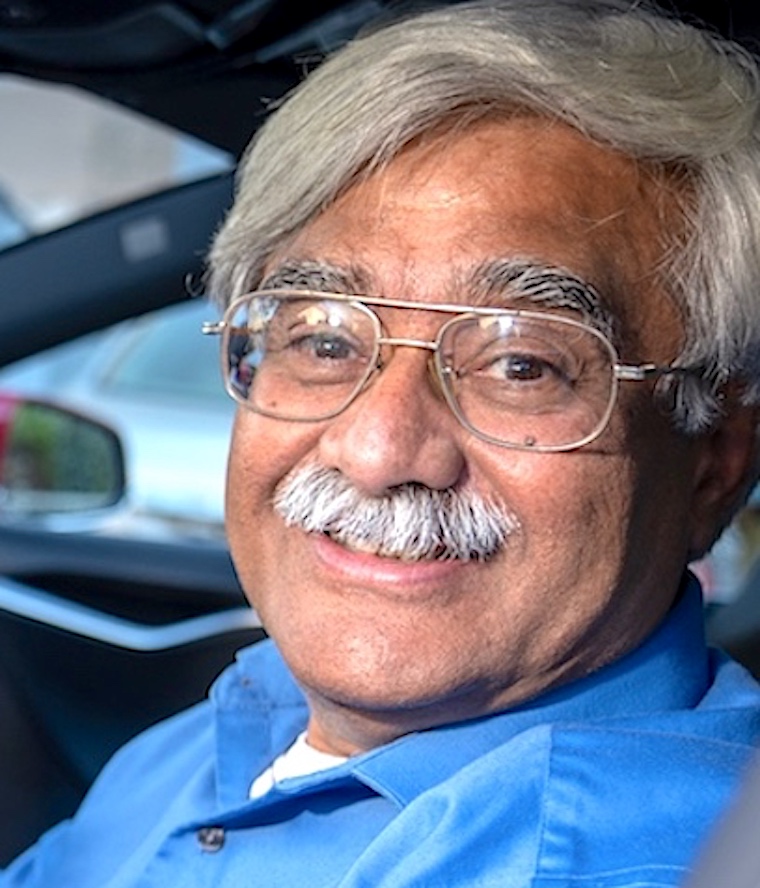}}]{Mohan Manubhai Trivedi}
is a Distinguished Professor at University of California, San Diego (UCSD) and the founding director of the UCSD LISA: Laboratory for Intelligent and Safe Automobiles,
winner of the IEEE ITSS Lead Institution Award (2015). Currently, Trivedi and his team
are pursuing research in intelligent vehicles, machine perception, machine learning, human-robot interactivity, driver assistance, active safety systems. Three of his students have received ``best dissertation" recognitions. Trivedi is a Fellow of IEEE, ICPR and SPIE. He received the IEEE ITS Society's highest accolade ``Outstanding Research Award" in 2013. Trivedi serves frequently as a consultant to industry and government agencies in the USA and abroad. 
\end{IEEEbiography}

\end{document}